%% file: main.tex
\documentclass{article}
\usepackage{ijcai22}
\usepackage{xcolor}
\usepackage{times}
\usepackage{soul}
\usepackage{url}
\usepackage[hidelinks]{hyperref}
\usepackage[utf8]{inputenc}
\usepackage[small]{caption}
\usepackage{graphicx}
\usepackage{amsmath}
\usepackage{amsthm}
\usepackage{booktabs}
\usepackage{algorithm}
\usepackage{algorithmic}
\usepackage{tikz}
\usetikzlibrary{arrows.meta}
\usetikzlibrary{shapes}

\usepackage{amssymb}
\usepackage{multirow}
\usepackage{pifont}%
\newcommand{\cmark}{\ding{51}}%
\newcommand{\xmark}{\ding{55}}%

\newcommand{\mc}[1]{\mathcal{#1}}

\newcommand{\fish}{\textit{Fish}}
\newcommand{\shark}{\textit{Shark}}
\newcommand{\trout}{\textit{Trout}}

\title{Deep Learning with Logical Constraints} %
\author{Eleonora Giunchiglia$^{1}$\and Mihaela Catalina Stoian$^{1}$\And Thomas Lukasiewicz$^{2,1}$
\affiliations
$^1$\,Department of Computer Science, University of Oxford, UK \\
$^2$\,Institute of Logic and Computation, TU Wien, Austria \\
\emails
firstname.lastname@cs.ox.ac.uk}

\begin{document}
\maketitle

\begin{abstract}
    In recent years, there has been an increasing interest in  exploiting logically specified background knowledge in order to obtain neural models (i) with a better performance, (ii) able to learn from less data, and/or (iii) guaranteed to be compliant with the background knowledge itself, e.g., for safety-critical applications. In this survey, we retrace such works and categorize them based on (i) the logical language that they use to express the background knowledge and (ii)~the goals that they achieve. 
    
\end{abstract}

\section{Introduction}

The recent history of deep learning is a story of successes (see, e.g., \cite{senior2020alphafold}). Neural networks have been increasingly applied to solve everyday problems, often successfully, but  sometimes also showing  their potential flaws (see, e.g., \cite{wexler2017}). A major source of such shortcomings is that neural networks are still  domain-agnostic in most cases, as they often ignore domain knowledge about the problem at hand 
\cite{stevens2020AIforscience}. For this reason, in recent years, there has been an increasing interest in incorporating background knowledge into deep learning algorithms. Such background knowledge can be expressed in many different ways (e.g., algebraic equations, logical constraints, and natural language) and incorporated in neural networks   (i) to improve their performance (see, e.g., \cite{li2019namedneurons}),  (ii) to make them able to learn from less data (see, e.g., \cite{xu2018semanticloss}), and/or (iii) to guarantee that their behavior is compliant with the background knowledge itself (see, e.g., \cite{hoernle2022multiplexnet}). 
We remark the importance of the last point when considering safety-critical applications in which the background knowledge corresponds to requirements on the models. Indeed, we envision that, in the future, a requirement specification will become a
standard step in the development of machine learning models, as it is in any software development process. 

In this survey, we 
conduct a comprehensive and fine-grai\-ned analysis of the works in which  background knowledge is  expressed as constraints in a logic-based language and then exploited to obtain better models. We organize the papers into four macro-categories based on the richness of the language that they use to express the constraints. 
Furthermore, to maximize the usefulness of this survey, we take a 
problem-oriented approach, and for each considered paper, we report which shortcomings the authors aim to address with the inclusion of the logical background knowledge.  A summary of our categorization is given in Table~\ref{tab:model_list}. 

The rest of this survey is organized as follows. In Section~\ref{sec:def}, we first give a formal definition of the problem of learning with logical constraints. Sections \ref{sec:binary} to \ref{sec:fol} then describe the models that belong to each macro-category,  
starting from the one with the simplest corresponding language, up to the one with the richest. In Section \ref{sec:concl}, we end the paper with some concluding remarks and pointers to other related surveys.

\section{Learning with Logical Constraints}\label{sec:def}

We formalize the problem 
of {\sl learning with logical constraints} as a triple $\mc{P} = (\mc{C}, \mc{X}, \Pi)$:
\begin{enumerate}
    \item $\mc{C}$ is a pair $(\mc{I},\mc{O})$, where $\mc{I} = I_1, I_2, \ldots, I_d$ ($d \geq 1$) are the {\sl input features}, and $\mc{O} = O_1, O_2, \ldots, O_n$ ($n \geq 1$) are the {\sl outputs}. Each input feature $I$ (resp., output $O$) is associated with a non-empty domain $D_I$ (resp., $D_O$) of {\sl values}, and $I$ (resp., $O$) is {\sl Boolean} when $D_I = \{0,1\}$ (resp., $D_O=\{0,1\}$). $D_\mc{I} = D_{I_1} \times \ldots \times D_{I_d}$ (resp., $D_\mc{O} = D_{O_1} \times \ldots \times D_{O_n}$) is the set of the possible inputs (resp., outputs). A {\sl data point} is an element of $D_\mc{I}$.
    \item
    $\mc{X}$ is a finite set of pairs $(x,y)$, where $x$ is a (possibly partially specified) data point, and 
    $y$ is its (possibly partially specified) {\sl ground truth}.
    \item
    $\Pi$ is a finite non-empty set of first-order constraints, which   delimit the set of outputs that can be meaningfully associated with  each data point. We assume that the constraints are written using  (i)
    for each Boolean (resp., non-Boolean) input feature $I$ a corresponding $d$-ary predicate (resp., function) $I$: intuitively, $I(x)$ (resp., $I(x) = z$) means that the value of the input feature $I$ in the data point $x$ is 1 (resp., $z$), and analogously for each Boolean (resp., non-Boolean) output $O$,
    (ii) variables ranging over data points,  (iii) specific values, and (iv) possibly other logical symbols (like ``$\neg$", ``$\wedge$", ``$\vee$", ``$\to$", ``$\forall$", ``$\exists$", ``$=$", ``$\geq$", and  ``$+$"). To simplify the formal treatment and save space, we assume that each constraint is closed and in prenex form (i.e., that each variable is either existentially or universally quantified at the beginning of the constraint).
    \end{enumerate}
    Notice that our definitions of both $\mc{X}$ and $\Pi$ are very general. This is necessary (i) for $\mc{X}$,  to cover papers dealing with supervised and semisupervised problems, and (ii) for $\Pi$, given that each paper uses its own specific language to model the learning problem and exploit the background knowledge. 
    For example, consider the classification problem associated with CIFAR-100 \cite{cifar100}. 
If we associate with each class and superclass a separate Boolean output, it is possible to formalize the
knowledge that sharks and trouts are fish via 
\begin{equation}\label{eq:shark}
\forall x. (\shark(x) \to \fish(x)), \ \  \forall x. (\trout(x) \to \fish(x)),
\end{equation}
    and that sharks are not trouts via 
    \begin{equation}\label{eq:fish}
            \forall x. (\shark(x) \to \neg \trout(x)).
    \end{equation}
In the above formulas, $\shark$, %
$\fish$, and $\trout$ are  predicates, each corresponding to a Boolean output of the network. Notice that the above constraints can be equivalently written in many different ways, exploiting  well-known first-order and propositional logic equivalences. Still, in the literature, constraints are often written and handled as  {\sl rules}, i.e., formulas %
like (\ref{eq:shark}) and (\ref{eq:fish}), having one of the following two forms
\begin{equation}
    \label{eq:r}
\forall x. (F(x) \to A(x)), \qquad
\forall x. (F(x) \to \neg A(x)), 
\end{equation}
where $F(x)$ is a conjunction of {\sl literals} (i.e., atomic formulas and negations of atomic formulas, where a formula is {\sl atomic} if it does not contain propositional connectives and/or quantifiers), and $A$ is a predicate associated with one Boolean output.
The motivation of  introducing rules as a special case of formulas is based on the fact that most of the works express  background knowledge as rules of the form (\ref{eq:r}), and then, for each data point $x$, use the value computed for 
$F(x)$ to determine the value to be associated with $A(x)$. Furthermore, in the literature, there has been a special focus on basic rules in which also $F$ (like $A$) is a predicate associated with a Boolean output. Rules in (\ref{eq:shark}) and (\ref{eq:fish}) are basic. 

We first survey papers in which constraints are basic rules (Section \ref{sec:binary}) and then the ones allowing for general rules (Section \ref{sec:normal}).
All the other surveyed papers are then divided into two categories, depending on whether they allow only for universal quantification (Section \ref{sec:prop}) or also for existential quantification (Section \ref{sec:fol}) in the constraints.
Thus, the fragment of first-order logic considered in each section allows for a richer language than the one allowed in the preceding sections.

\section{Basic Rules} \label{sec:binary}

Basic rules are expressions of the form (\ref{eq:r}) where both $F$ and $A$ are predicates corresponding to Boolean outputs. 
They have been used in the context of multi-label classification (MC) problems with constraints, being the special case of learning with constraints in which all the outputs are Boolean labels. Thus, each output has a corresponding associated predicate, and the rules specify the existing relations between the output labels. %
All reviewed papers use rules of the first form in (\ref{eq:r}) to express a hierarchical relation between two output labels, and some papers use also rules of the second form in (\ref{eq:r}) to express a mutual exclusion between two output labels.
Thus, we further divide the papers depending on their usage of basic rules of the second form in (\ref{eq:r}).

\paragraph{Hierarchical rules.} 
In learning problems with hierarchical rules, knowledge is expressed with basic rules of the first form in (\ref{eq:r}) in which also $F$ (like $A$) is a Boolean output. 
Examples of hierarchical rules are (\ref{eq:shark}), which are satisfied whenever a data point $x$ predicted to be a shark or a trout is also predicted to be a fish. 
Interestingly, an entire field has been developed to deal with such rules, namely, {\sl hierarchical multi-label classification} (HMC) problems, which are MC problems with hierarchical rules satisfying the additional assumption that there are no cycles between the labels, when drawing an arc from the antecedent to the consequent of each rule (see, e.g., \cite{cerri2011}). %
HMC problems naturally arise in many different domains, such as image classification or functional genomics, and HMC models have normally two goals: (i)~improve on the SOTA models, and (ii)~guarantee the satisfaction of the hierarchical constraints. As expected, many different neural models were developed for HMC problems. 
To present them, we follow the classic categorization used for general HMC models, which divides them  into two groups based on  how they exploit the hierarchical knowledge \cite{silla2011}: 
\begin{enumerate}
\item {\sl local approaches} exploit the constraints to decompose the problem into smaller classification ones, and then combine the solutions appropriately, while
\item {\sl global approaches}  consist of single models able to associate  objects with their corresponding classes in the hierarchy as a whole. 
\end{enumerate}
 Local approaches  can be further divided into three subcategories based on the strategy that they deploy to decompose the main task. The most popular strategy is {\sl local classifier per level}, which is given when  a method trains a different classifier for each level of the hierarchy. For example, HMC-LMLP \cite{cerri2011,cerri2014} is a model consisting of a multi-layer perceptron (MLP) per hierarchical level, and the predictions in one level are used as inputs to the network responsible for the predictions in the next level. This model was later extended in \cite{cerri2018}, where HMCN is proposed. HMCN is considered a hybrid model, because it is trained with both a local and a global loss.
Finally, DEEPre \cite{li2018hmc}, later extended in mlDEEPre \cite{zou2019hmc}, consists of a neural network %
for each level of the hierarchy, and it is applied to the challenging problem of enzyme function prediction. On the other hand, if a method trains a classifier for each node of the hierarchy, then we have a {\sl local classifier per node}. An example model for this category is  HMC-MLPN \cite{feng2018}, in which one MLP for each node is deployed. Finally, if a method trains a different classifier per parent node in the hierarchy, then we have a {\sl local classifier per parent node}. For example, \citeauthor{kulmanov2018} \shortcite{kulmanov2018} propose DeepGO, in which   a small neural-based model for each subontology of the Gene Ontology  is trained. %
Global methods, on the other hand, do not have any subclassification, and global neural models are quite recent. The first proposed global model based on neural networks is AWX \cite{masera2018}, which is just a feed-forward neural network predicting the leaves of the hierarchy, and then inferring the value for the parent nodes. A more complex model is given by C-HMCNN \cite{giunchiglia2020}, which builds a constraint layer ensuring the satisfaction of the constraints. Such a layer works synergistically with a constraint loss to exploit the background knowledge of the hierarchy. Finally, MBM \cite{patel2022MBM} takes an alternative approach by representing the labels by boxes rather than vectors, and thus it is able to capture taxonomic relations among labels.

\begin{figure}
    \centering
   \resizebox{0.75\linewidth}{!}{\hspace{0cm}\input{tikz/newbinary.tikz}\hspace{0cm}}
    \caption{
    Example of a hierarchy DAG (left) and of a hierarchy and exclusion graph (right). Here, ``$\to$'' indicates a hierarchical relation, while ``{\color{blue} \textbf{---}}'' indicates mutual exclusion.\vspace*{-2ex}}
    \label{fig:dag_hex}
\end{figure}
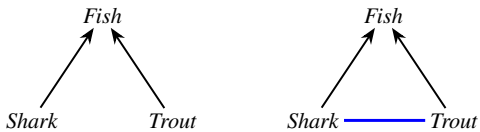

\begin{table*}[ht!]
\setlength{\tabcolsep}{5.2pt}
    \centering
    \footnotesize
    \begin{tabular}{l l c c}
    \toprule
    \textbf{Example Rules} & \textbf{Common High-Level Algorithm} & \textbf{KBANN Mapping} & \textbf{CIL$^2$P Mapping} \\
    \midrule 
        $\forall x.(A_1(x) \wedge \neg A_2(x) \to A_4(x))$ & 1. Map rules to the network. & \multirow{7}{*}{\resizebox{0.13\linewidth}{!}{\centering\hspace{-0.1cm}\input{tikz/kbann.tikz}\hspace{0.0cm}}} & \multirow{7}{*}{\resizebox{0.275\linewidth}{!}{\centering\hspace{-0.0cm}\input{tikz/cilp.tikz}\hspace{0.0cm}}}\\
        $\forall x.(A_3(x) \wedge A_4(x) \to A_5(x))$ & 2. Add $n$ and $m$ neurons to the input \\
        {\color{blue}{$\forall x.(A_5(x) \to A_4(x)) $}} &  \quad and hidden layers, respectively. \\
        & \quad $n,m$ are user-defined parameters. \\
        & 3. Fully connect the newly added \\
        & \quad neurons to the adjacent layers.\\
        & 4. Perturb all the network's weights. \\
    \bottomrule
    \end{tabular}
    \caption{KBANN and CIL$^2$P comparison. Given the sample rules (first column), we provide the high-level algorithm that the two models share to build the networks from the rules (second column), and then we show how the rule mapping (Step 1) is done in KBANN (third column) and CIL$^2$P (fourth column), respectively. We highlighted {\color{blue} $\forall x.(A_5(x) \to A_4(x))$}, because while CIL$^2$P can map it in the network, KBANN must exclude it in order to mantain the acyclicity of the rules ($\forall x.(A_5(x) \to A_4(x))$ is indeed absent in KBANN).\vspace*{-2ex}}
    \label{tab:kbann_cilp_comparison}
\end{table*}

\paragraph{Hierarchical and exclusion rules.} In computer vision, hierarchical rules are often used together with exclusion constraints expressed as basic rules of the second form in (\ref{eq:r}) in which $F$ (like $A$) is a Boolean output. An example of a basic rule corresponding to an exclusion constraint is (\ref{eq:fish}), which is satisfied whenever a data point $x$ predicted to be a shark cannot be predicted to also be a trout. 
Hierarchical and mutual exclusion rules can be represented together in a hierarchy and exclusion (HEX) graph, firstly proposed in \cite{deng2014hexgraphs}. An example of a HEX graph is given in Figure~\ref{fig:dag_hex}. In addition to proposing HEX graphs, \citeauthor{deng2014hexgraphs} also build a neural model guaranteed to satisfy the rules and able to deal with incomplete labels. HEX graphs are often applied to tackle the problem of fine-grained image classification. Fine-grained image classification  refers to the problem of classifying images 
 that possess very subtle discriminatory features, e.g., classifying images of different birds by species, or of  flowers by categories. Their usage in the field was firstly proposed in \cite{xie2015_binary_class_add},
 where the authors tackle the problem 
by adding a set of classes representing broader concepts than the initial ones (all the newly added classes are thus parent nodes in the hierarchy), and then acquiring a large number of images from external sources, which are labelled with such newly added classes. The new task is thus to predict both the initial label and the newly annotated labels. 
The authors show how, due to the added data,
they achieve better results on this new task. An even more interesting study is shown in \cite{chang2021flamingo}, where the authors  
show that it is possible to exploit the background knowledge to build a model that is able to get better results than the SOTA models on the initial set of classes.  \citeauthor{chang2021flamingo}'s results have been recently surpassed by HRN \cite{chen2022labelhex}, a network able to perform hierarchical feature interaction via residual connections.

\section{General Rules} \label{sec:normal}

In this section, we lift the assumption that $F(x)$ in rules of the type (\ref{eq:r})  is a single literal, and we review the papers in which $F(x)$ is allowed to be a conjunction of $m$ literals $(m \geq 1)$. We classify such papers depending on whether $F(x)$ can (or cannot) contain  functions and/or predicates corresponding to input features. In the first case, given a data point $x$, the rules constrain the output label 
$A$ also on the basis of the input $x$, and thus each rule corresponds to an input-output constraint. In the second case, each rule excludes some output configurations independently from the data point $x$.

\paragraph{Input-output constraints.} Interestingly, this type of constraints was the very first to be studied in this field. This is probably due to the fact that before the spread of machine learning systems, many classifiers were still hand-built, and thus researchers had a large availability of rules of the type ``if the input has these features, then the data point belongs to this class''. The goal of these models was thus to overcome the flaws of both hand-built classifiers and learned classifiers (which could be seen as almost complementary flaws) by creating hybrid systems. One of the first works able to incorporate such rules in the topology of the neural network was KBANN \cite{shavlik1989kbann_aaai,towell1994kbann}. Given a set of rules, \citeauthor{shavlik1989kbann_aaai} \shortcite{shavlik1989kbann_aaai} 
map the supporting facts to the input neurons, the intermediate conclusions to the hidden neurons, and the final conclusions to the output neurons. Given such a mapping, clearly, KBANN needs the additional assumption that the constraints are acyclic. KBANN was later extended by \citeauthor{fu1993kbcnn} \shortcite{fu1993kbcnn}, who proposed KBCNN. Given a set of acyclic rules, not only KBCNN is able to map the rules into a neural network like KBANN, but also it is able to map back the neural network to a set of learned rules. Such learned rules have the advantage of being completely transparent to the human user. While both KBANN and KBCNN build a neural network directly from the rules, in \cite{obradovic1993hidneuronsadd}, the rules are treated as an expert system that is refined through incremental hidden unit generation, which allows the model to learn new rules without corrupting the initial ones (as often happens in KBANN), and to get a good performance even with severely incomplete rule bases (contrarily to KBANN). Another model proposed to solve such problems is Cascade ARTMAP~\cite{Tan1997cascade_artmap}, which is able to learn a set of rules that are more accurate and simpler than the ones extracted from KBANN. While all the methods explored so far can only deal with acyclic rules, one of the first methods able to deal with cyclic rules was CIL$^2$P  \cite{garcez1999cil2p}. CIL$^2$P and KBANN share the same high-level learning algorithm, however, they map the rules into neural networks in different ways. Due to such a different mapping, not only \citeauthor{garcez1999cil2p} can map cyclic rules, but they are also able to prove that CIL$^2$P computes the stable model of the logic program represented by the neural network, thus making it a  parallel system for logic programming. 
Notice that CIL$^2$P was later extended in \cite{franca2014cilp++} to represent and learn  first-order logic formulas. This new system is called CILP++.
Since KBANN and CIL$^2$P are probably the most well-known early models able to combine logic and neural networks, we provide a more detailed comparison between them in Table~\ref{tab:kbann_cilp_comparison}. 
More details on the approaches from the 90s to combine  knowledge engineering and machine learning are given in the surveys \cite{shavlik1994oldmethods_oldsurvey,shavlik2021oldmethods_survey}.

\paragraph{Constraints over the output.} While all these early models focus on input-output constraints, the more recent ones focus on constraints over the output domain. From a high-level perspective, we can further divide these models based on how they integrate the neural network with the constraints. In the first set of models,  learning and reasoning represent two different steps in a pipeline, which is though trained end-to-end. On the other hand, in the second set of models, the constraints are directly integrated in the network structure and/or loss function, thus we have ``single-stage models''.

\smallskip\noindent  
\emph{Pipelines:} Probably one of the most famous methods able to constrain neural networks' outputs is DeepProbLog~\cite{manhaeve2018deepproblog}. DeepProbLog extends  ProbLog~\cite{deraedt2007problog} by encapsulating the outputs of the neural network in the form of neural predicates. This can be easily done,  because (in ProbLog) atomic expressions are already assigned a probability. Furthermore, since the algebraic extension of ProbLog \cite{kimmig2011problog_extension} already supports automatic differentiation, it is possible to backpropagate the gradient from the loss at the output through the neural predicates into the neural networks, thus allowing the neural networks to learn directly from the logical constraints.  DeepProbLog can thus be seen as a two-stage pipeline, where the neural networks handle the low-level perception, and then the reasoning is done at a logical level. Very close to DeepProbLog is NeurASP \cite{Yang2020NeurASP}, however, differently from DeepPobLog, NeurASP employs reasoning originating from answer set programming, such as defaults, aggregates, and optimization rules. A deeper pipeline, in which each stage models a sub-task of a more complex problem, has been proposed in \cite{sachan2018nuts_and_bolts}, where the authors present Nuts\&Bolts: a framework in which each stage can be a set of different function approximators, and that can be trained end-to-end to minimize a global loss function. Nuts\&Bolts is able to exploit the background knowledge by expressing it as rules, then translating them into probabilistic soft logic and incorporating it as one of the function approximators. Notice that, in this case, the rules do not only express constraints over the final outputs of the pipeline, but also over the outputs of the intermediate stages.

\smallskip\noindent  
\emph{Single-stage models:} Among these models, we find CNN, a model proposed in \cite{giunchiglia2021}. In this work, the authors present a way to map the constraints (i) into a top-layer (which can be built on top of any neural network), and (ii) into a loss function (which is able to work synergistically with the layer to exploit the background knowledge and get better results). Thanks to the top layer, \citeauthor{giunchiglia2021} are able to guarantee that the constraints are always satisfied, however, they also make the assumption that the set of constraints must be stratified. \citeauthor{li2019namedneurons} \shortcite{li2019namedneurons} also change the structure of neural networks to incorporate the background knowledge expressed by the constraints. To this end, the authors recognize that, in a neural network, some neurons (called {\sl named neurons}) can be endowed with semantics tied to the task, and that logical rules can be written over such named neurons. Thanks to this particular formulation, \citeauthor{li2019namedneurons} are able to inject constraints over all the outputs of the neural network, even the intermediate ones (e.g., the outputs of the attention layer). The constraints that they are able to capture are thus over (i)~the output domain, (ii) the intermediate outputs, and (iii) the relations between the intermediate outputs and the final outputs. \citeauthor{minervini2018adversarial}  \shortcite{minervini2018adversarial}, on the other hand, map the constraints to a loss function that is used to generate adversarial examples that cause a model to violate preexisting background knowledge. In particular, they apply their method to the task of recognizing textual entailment, and they add rules like ``if sentence $s_1$ contradicts sentence $s_2$, then $s_2$ contradicts $s_1$ as well", ultimately showing that their models obtain a much better performance on adversarial datasets.

\begin{table*}[ht!]
    \centering
    \resizebox{0.78\textwidth}{!}{%
    \begin{tabular}{lcccc}
    \toprule
       \textbf{Model}  & \textbf{Expressivity} & \textbf{\begin{tabular}[c]{@{}c@{}}Guaranteed\\ Satisfaction \end{tabular}}  & \textbf{\begin{tabular}[c]{@{}c@{}}Less Data/\\ Supervision \end{tabular}} & \textbf{\begin{tabular}[c]{@{}c@{}}Improve \\on SOTA \end{tabular}}  \\
       \midrule 
        HMC-LMLP~\cite{cerri2011} &  \multirow{10}{*}{basic rules} &   \xmark & \xmark & \cmark \\
        HEX~\cite{deng2014hexgraphs} &  & \xmark & \cmark & \cmark \\
        HAR-CNN~\cite{xie2015_binary_class_add} &  & \cmark & \xmark & \cmark \\
        DEEPre~\cite{li2018hmc}&  &  \cmark & \xmark & \cmark\\
        HMCN~\cite{cerri2018}& & \xmark & \xmark & \cmark \\
        AWX~\cite{masera2018}& & \cmark &  \xmark & \cmark \\
        DeepGO~\cite{kulmanov2018}& & \cmark & \xmark & \cmark\\
        mlDEEPre~\cite{zou2019hmc}& & \cmark & \xmark & \cmark \\
        C-HMCNN~\cite{giunchiglia2020}&  & \cmark & \xmark & \cmark \\
        MBM~\cite{patel2022MBM}& & \xmark & \xmark & \cmark\\
       \midrule
         KBANN~\cite{shavlik1989kbann_aaai}  & \multirow{10}{*}{general rules} &  \xmark &\xmark & \cmark \\
        KBCNN~\cite{fu1993kbcnn} & & \xmark &\xmark & \cmark  \\
        Iterative Neurons Addition~\cite{obradovic1993hidneuronsadd}& & \xmark &\xmark & \cmark \\
        Cascade ARTMAP~\cite{Tan1997cascade_artmap}& & \xmark &\xmark & \cmark \\ 
        CIL$^2$P~\cite{garcez1999cil2p}& & \xmark &\xmark & \cmark \\
        Adversarial Regularisation~\cite{minervini2018adversarial} & &  \xmark & \xmark & \cmark \\
        DeepProbLog~\cite{manhaeve2018deepproblog} & &  \xmark & \cmark & \xmark \\
        NeurASP~\cite{Yang2020NeurASP} & &  \xmark & \cmark & \xmark \\
        CCN~\cite{giunchiglia2021} & &  \cmark &\xmark &\cmark \\
        Nuts\&Bolts~\cite{sachan2018nuts_and_bolts}  & &   \xmark & \cmark  & \cmark \\
        Named Neurons~\cite{li2019namedneurons} & & \xmark & \cmark & \cmark \\
        \midrule
        Label-free Supervision~\cite{stewart2017ermon} &  \multirow{6}{*}{{\begin{tabular}[c]{@{}c@{}}universally\\ quantified formulas \end{tabular}}} & \xmark & \cmark & \xmark \\
        Semantic \ Loss~\cite{xu2018semanticloss} & & \xmark & \cmark & \xmark \\
        LENSR~\cite{xie2019lensr} &  & \xmark & \xmark & \cmark\\
        DL2~\cite{fischer2019DL2} &  & \xmark & \cmark & \xmark\\
        NESTER~\cite{dragone2021nester} &  & \cmark & \cmark & \xmark \\
        MultiPlexNet~\cite{hoernle2022multiplexnet} & & \cmark & \cmark & \xmark \\
        \midrule 
        SBR~\cite{diligenti2012SBR} & \multirow{9}{*}{\begin{tabular}[c]{@{}c@{}}universally and  existentially\\ quantified formulas \end{tabular}} & \xmark & \cmark & \xmark\\
        CILP++~\cite{franca2014cilp++} & & \xmark & \xmark & \cmark \\
        LTN~\cite{serafini2016ltn} & &  \xmark & \xmark &\cmark \\
        Iterative Rule Distillation~\cite{hu2016harnessing} &  & \xmark & \cmark & \cmark\\
        Mutual Iterative Rule  Distillation~\cite{hu2016mutual}  & & \xmark & \cmark & \cmark\\
        LTN-SII~\cite{donadello2017} &  & \xmark & \xmark &\cmark \\
        LYRICS~\cite{marra2019lyrics} &  & \xmark & \cmark & \xmark \\
        ABL~\cite{dai2019abductive} &  & \xmark & \cmark & \xmark\\
        DFL~\cite{krieken2020dfl} &  & \xmark & \cmark & \xmark \\
    \bottomrule
    \end{tabular}}
    \caption{Summary table of the analyzed works. For each work, we report which problem the authors wanted to tackle through the inclusion of the constraints. We identified three common problems tackled in the literature: (i) building models whose outputs are guaranteed to be compliant with a set of constraints, (ii) exploiting the background knowledge to get a better performance in either low-data regimen settings or to  exploit unlabelled data, and (iii) beating SOTA models on benchmark datasets (Columns 3 to 5). If a model exploits the background knowledge for one of the three tasks above, then it is marked with ``\cmark'', and with ``\xmark{}'', otherwise.\vspace*{-2.5ex}}
    \label{tab:model_list}
\end{table*}

\section{Universally Quantified Formulas} \label{sec:prop}
 
In this section, we describe methods that allow for constraints expressed as  universally quantified first-order formulas. All the methods belonging to this category either inject the constraints in the loss (loss-based methods) or they constrain the value of the outputs (constrained-output methods).

\paragraph{Loss-based methods.} While these methods obviously cannot guarantee the satisfaction of the constraints, they are able to generalize from less data and/or exploit unlabelled data.  
One of the first papers to propose to exploit background knowledge to solve the problem in data-scarce settings was \cite{stewart2017ermon}. In this paper, the authors show how it is possible to train neural networks to detect objects without any direct supervision, and by simply injecting in the loss the constraints on the output domain. In their work, they considered both  constraints pertaining to dynamics equations and logical constraints. 
However, in each of the proposed examples, the authors had to manually engineer the loss function. After this work, a number of papers on how to automatically map universally quantified constraints into a loss function were studied.
For example, in \cite{xu2018semanticloss}, a {\sl semantic loss} is proposed. Given a data point and a neural network that outputs the vector of probabilities $\pmb{p}$,  the semantic loss is defined such that it is proportional to the negative logarithm of the probability of generating an interpretation that satisfies the constraints, when sampling values according to~$\pmb{p}$. This method has the great quality of being completely syntax-independent (i.e., no matter how we write the constraints, the value of the loss does not change). 
Another model that incorporates the logical constraints in the loss is LENSR~\cite{xie2019lensr}. However, in this model, the formulas are first rewritten in either conjunctive normal form or decision deterministic decomposable negation normal form, and then projected onto a manifold where entailment is related to distance. Once learned,  the logic embeddings can then be used to form a logic loss that guides the neural network training by encouraging formula embeddings to be close to satisfying assignments, and far from unsatisfying assignments. Yet another method that translates the constraints in continuous terms to be added to the loss is DL2 \cite{fischer2019DL2}. What is particularly interesting about this method is that it allows for queries. For example, a user can inquire the models on which neurons took part in a decision, or to generate adversarial examples violating a given constraint. Furthermore, notice that the language includes  Boolean combinations of comparisons between terms, where a term is any real-valued function that may appear as a subexpression of a loss function. 

\paragraph{Constrained-output methods.} The methods presented here have the goal of guaranteeing the satisfaction of the constraints, and thus they apply the constraints directly on the outputs. MultiPlexNet~\cite{hoernle2022multiplexnet} further extends the language,  as its constraints can consist of any quantifier-free linear arithmetic formula over the rationals (thus, involving ``$+$", ``$\ge$", ``$\neg$", ``$\wedge$", and ``$\vee$"). 
In MultiPlexNet, the formulas are expressed in disjunctive normal form (DNF), and then the output layer of an existing neural network is augmented to include a separate transformation for each term in the DNF formula. Thus, the network’s output layer can be viewed as a multiplexor in a logical circuit that permits for a branching of logic, and exactly from this property, the model gets the name MultiPlexNet. Another work that limits the output space of the neural network is NESTER \cite{dragone2021nester}. However, in this case, the constraints are not mapped into the last layer of the  network (like MultiPlexNet or CNN), but they are enforced by passing the outputs of the neural network to a constraint program, which enforces the constraints. NESTER can enforce hard and soft constraints  over both categorical and numerical variables, and the entire architecture can be trained end-to-end by backpropagation.

\section{Universally and Existentially Quantified Formulas} \label{sec:fol}

Our final category contains the models that allow for both universal and existential quantification. We divide the works between those that inject the constraints into the loss (loss-based methods) and those that create specialized neural structures to incorporate the constraints into the topology of the neural network (specialized structure-based methods).

\paragraph{Loss-based methods.} One of the first models developed to this end was {\sl semantic-based regularization} (SBR) \cite{diligenti2012SBR,diligenti2017sbrnn}. Like the above loss-based methods, SBR was developed with the goal of exploiting unlabelled data to train machine learning models. In this work, the authors map the constraints to a differentiable regularization term that can be added to any loss. To this end, they perform the following steps: (i) the first-order logic (FOL) expressions are grounded, (ii) the quantifier-free formulas are mapped to real-valued functions using the t-norms~\cite{klement2000}, and (iii) each FOL formula containing a universally (resp., existentially) quantified variable $x$ is mapped to a function that returns the average (resp., maximum) of the t-norm generalization when grounding $x$ over its domain. Other methods take inspiration from the distillation method proposed in \cite{hinton2015distilling}, and have the goal of creating models able to both beat the SOTA models and exploit unlabelled data.  In \cite{hu2016harnessing}, the authors propose an {\sl iterative rule knowledge distillation} procedure that transfers the structured information encoded in the logical rules into the network. In particular, at each iteration, a rule-regularized neural teacher is built, and then a student network is trained to imitate the predictions of the teacher network, while also trying to predict the right label for as many data points as possible. This work was later extended in \cite{hu2016mutual}. This framework iteratively transfers information between the neural network and the structured knowledge, thus resulting in an effective integration of the representation learning capacity of neural networks and the generalization power of the structured knowledge. 
Different from distillation-based methods, but following a similar reasoning about how unifying two complementary paradigms can yield mutual benefits and surpass SOTA models, an abductive learning (ABL) approach is introduced in \cite{dai2019abductive}.
ABL combines a neural network, responsible for interpreting subsymbolic data into primitive logical facts, with a logical component able to reason on these facts. 
If the generated facts are inconsistent with the background knowledge, ABL generates new pseudo-labels that more likely satisfy the background knowledge and retrains the neural network \mbox{accordingly.} %

\paragraph{Specialized structure-based methods.} Instead of mapping the constraints into the loss function, \citeauthor{serafini2016ltn} \shortcite{serafini2016ltn} and  \citeauthor{serafini2022ltn} \shortcite{serafini2022ltn} map functions and predicates to matrices, and constants to vectors. They propose real logic: a logical formalism on a first-order language whereby formulas have a truth value in the interval [0,1] and a semantics defined concretely on the domain of real numbers. They then show how real logic can be encoded in logic tensor networks (LTNs). \citeauthor{donadello2017}~\shortcite{donadello2017} later show how to use LTNs for semantic image interpretation, and \citeauthor{marra2019lyrics}~\shortcite{marra2019lyrics} present LYRICS, an extension of LTNs that provides an input language allowing for   background knowledge in FOL, where predicate and function symbols are grounded onto any computational graph. All the logics  in the works above belong to the family of differentiable fuzzy logics (DFL) of  \cite{krieken2020dfl,krieken2022dfl}. DFL is a family of differentiable logics, where the term {\sl differentiable logics} refers to a logic along with a translation scheme from logical expressions to differentiable loss functions. So, differentiable fuzzy logics stand for the case where the logic is a fuzzy logic and the translation scheme applies to logical expressions that  include fuzzy operators. Van Krieken \emph{et al.}\nocite{krieken2020dfl} analyze DFL's behavior over different choices of fuzzy logic operators and give some recommendations on the choice of the operators.

\vspace*{-0.5ex}
\section{Summary and Outlook}\label{sec:concl}

We conducted a comprehensive and fine-grained analysis of deep learning approaches in which  background knowledge is expressed and then exploited as logical constraints in first-order logic. We categorized the papers based on the richness of the logical language used to express the constraints. Furthermore, for each approach, we reported
which shortcomings the authors wanted to address via the background knowledge.
For coarser surveys on how to include background knowledge expressed in different ways, 
see~\cite{rueden2021informed,dash2022nature}. All the presented models  fall  into the broader field of neural-symbolic computing, whose aim is to integrate the abilities of learning and of reasoning about what has been learned. For a broad survey on neural-symbolic models see \cite{garcez2019survey}. 

We envision that the specification and exploitation of logical constraints in deep learning models will become more and more widespread in the future, especially in safety-critical applications, where requirements are often formally specified as logical formulas, thus combining the advantages of manually engineering safety-critical features and automatically learning all other features. One open challenge for future work thus includes an AI system engineering 
approach that gives strict guarantees in the form of logical constraints on the system behavior based on such a combined system design.

Another open challenge for future research concerns 
the exploitation of logical constraints in explainable AI. 
More precisely, logical constraints and reasoning about the  functionality
of neural networks and their output data  can actually be exploited in order to model the abstract reasoning that is underlying the predictions of these networks, and it can thus be used in order to naturally create explanations for these predictions. Only few existing papers, such as \cite{shih2020,DBLP:journals/corr/abs-2106-13876}, have started to explore this large potential of logical~constraints in deep learning to date in different ways.

\section*{Acknowledgments}
Eleonora Giunchiglia is supported by the EPSRC under the grant EP/N509711/1 and by an Oxford-DeepMind Graduate Scholarship. Mihaela Catalina Stoian is supported by the EPSRC under the grant EP/T517811/1. This work was also supported by the Alan Turing Institute under the EPSRC grant EP/N510129/1, by the AXA Research Fund, and by the EPSRC grant EP/R013667/1.

\bibliographystyle{named}
\bibliography{bibliography}
\end{document}

%% file: tikz/newbinary.tikz
\usetikzlibrary{arrows.meta}
\begin{tikzpicture}
\begin{scope}[every node/.style={rectangle,thick,draw=none,inner sep=0, outer sep=2}]
    \node (Fish) at (1,1) {\small{$\fish$}};
    \node (Shark) at (0,-0.5) {\small{$\shark$}};
    \node (Trout) at (2,-0.5) {\small{$\trout$}};
 
     \node (DogHEX) at (5,1) {\small{$\fish$}};
    \node (HuskyHEX) at (4,-0.5) {\small{$\shark$}};
    \node (BeagleHEX) at (6,-0.5) {\small{$\trout$}};   
\end{scope}
\begin{scope}[>={Stealth[black]},
              every edge/.style={draw=black, thick}]
     \path [<-] (Fish) edge node {} (Shark);
      \path [<-] (Fish) edge node {} (Trout);
      
      \path [<-] (DogHEX) edge node {} (HuskyHEX);
      \path [<-] (DogHEX) edge node {} (BeagleHEX);
\end{scope}

\begin{scope}[>={Stealth[black]},
              every edge/.style={draw=blue,very  thick}]
    \path [-] (HuskyHEX) edge node {} (BeagleHEX);
\end{scope}
\end{tikzpicture}

%% file: tikz/kbann.tikz
\usetikzlibrary{arrows.meta}
\begin{tikzpicture}
\centering
\begin{scope}[every node/.style={rectangle,thick,draw=none,inner sep=0, outer sep=2}]
    \node (A1) at (0,0) {$A_1$};
    \node (A2) at (1,0) {$A_2$};
    \node (A3) at (2,0) {$A_3$};
    
    \node (A4) at (0.5,1) {$A_4$};
    \node (A5) at (1,2) {$A_5$};
\end{scope}
\begin{scope}[>={Stealth[black]},
              every edge/.style={draw=black, thick}]
    \path [->] (A1) edge node {} (A4);
    \path [->] (A3) edge node {} (A5);
    \path [->] (A4) edge node {} (A5);
\end{scope}

\begin{scope}[>={Stealth[black]},
              every edge/.style={draw=black,thick,dashed}]
    \path [->] (A2) edge node {} (A4);
\end{scope}
\end{tikzpicture}

%% file: tikz/cilp.tikz
\usetikzlibrary{arrows.meta}
\begin{tikzpicture}
\centering
\begin{scope}[every node/.style={rectangle,thick,draw=none,inner sep=0, outer sep=2}]
    \node (A1) at (0,0) {$A_1$};
    \node (A2) at (1,0) {$A_2$};
    \node (A3) at (2,0) {$A_3$};
    \node (A4) at (3,0) {$A_4$};
    \node (A5) at (4,0) {\color{blue}{$A_5$}};
    
    \node (A4o) at (1.5,2) {$A_4$};
    \node (A5o) at (2.5,2) {$A_5$};
    
    \node (R1) at (1,1) {$R_1$};
    \node (R2) at (2,1) {$R_2$};
    \node (R3) at (3,1) {\color{blue}{$R_3$}};
\end{scope}

\begin{scope}[every node/.style={rectangle,thick,draw=none,inner sep=0, outer sep=0}]
    \node (A5ol) at (4.6,2) {};
     \node (A5l) at (4.6,0) {};
     
      \node (A4ol) at (-0.3,2) {};
      \node (A4l) at (-0.3,-0.5) {};
      \node (A4s) at (3,-0.5) {};
\end{scope}

\begin{scope}[>={Stealth[black]},
              every edge/.style={draw=black, thick}]
    \path [->] (A1) edge node {} (R1);
    \path [->] (A3) edge node {} (R2);
    \path [->] (A4) edge node {} (R2);
    
    \path [->] (R1) edge node {} (A4o);
    \path [->] (R2) edge node {} (A5o);

     \path [-] (A4o) edge node {} (A4ol);
    \path [-] (A4ol) edge node {} (A4l);
    \path [-] (A4l) edge node {} (A4s);
     \path [->] (A4s) edge node {} (A4);
\end{scope}

\begin{scope}[>={Stealth[black]},
              every edge/.style={draw=black,thick,dashed}]
    \path [->] (A2) edge node {} (R1);
\end{scope}

\begin{scope}[>={Stealth[blue]},
              every edge/.style={draw=blue,thick}]
    \path [->] (A5) edge node {} (R3);
    \path [->] (R3) edge node {} (A4o);
    
    \path [-] (A5o) edge node {} (A5ol);
    \path [-] (A5ol) edge node {} (A5l);
     \path [->] (A5l) edge node {} (A5);
\end{scope}

\end{tikzpicture}